\documentclass[runningheads]{llncs}
\usepackage{graphicx}
\usepackage{booktabs}

\usepackage{lipsum}
\usepackage{color}

\def\ie{{\em i.e.}}

\begin{document}
\title{LAE : Long-tailed Age Estimation}

\author{Zenghao Bao\inst{1,2,*} \and
Zichang Tan\inst{4,5,*} \and
Yu Zhu\inst{4,5} \and
Jun Wan\inst{1,2,\dagger} \and \\
Xibo Ma\inst{1,2} \and 
Zhen Lei\inst{1,2,3} \and
Guodong Guo\inst{4,5}
} 
\authorrunning{Bao et al.}

\institute{CBSR \& NLPR, Institute of Automation, Chinese Academy of Sciences \\ \email{ \{baozenghao2020, jun.wan\}@ia.ac.cn} \and
School of Artificial Intelligence, University of Chinese Academy of Sciences \\
\email{\{xibo.ma, zlei\}@nlpr.ia.ac.cn} \and
Centre for Artificial Intelligence and Robotics, 
Hong Kong Institute of Science \& Innovation, Chinese Academy of Sciences \and
Institute of Deep Learning, Baidu Research \and National Engineering Laboratory for Deep Learning Technology and Application \\
\email{\{tanzichang,zhuyu05,guoguodong01\}@baidu.com}}

\maketitle              
\newcommand\blfootnote[1]{%
\begingroup 
\renewcommand\thefootnote{}\footnote{#1}%
\addtocounter{footnote}{-1}%
\endgroup 
}
\blfootnote{*Co-First Author} 
\blfootnote{$\dagger$ Corresponding Author} 
\begin{abstract}

Facial age estimation is an important yet very challenging problem in computer vision. To improve the performance of facial age estimation, we first formulate a simple standard baseline and build a much strong one by collecting the tricks in pre-training, data augmentation, model architecture, and so on. Compared with the standard baseline, the proposed one significantly decreases the estimation errors. Moreover, long-tailed recognition has been an important topic in facial age datasets, where the samples often lack on the elderly and children. To train a balanced age estimator, we propose a two-stage training method named Long-tailed Age Estimation (LAE), which decouples the learning procedure into representation learning and classification. The effectiveness of our approach has been demonstrated on the dataset provided by organizers of Guess The Age Contest 2021.

\keywords{Deep Learning  \and Age Estimation \and Long-tailed Recognition.}
\end{abstract}
\section{Introduction}
As one of the most concerning topics in facial analysis, age estimation remains challenging for the uncontrolled nature of the aging process, a significant variance among faces in the same age range, and a high dependency of aging traits on a person. Since age estimation from face images is an inevitable problem in several real applications, enhancing age estimation performance has also become an urgent problem to be solved. In recent years, deep learning methods~\cite{niu2016ordinal,tan_pami_2018,gao2018age,tan2019deeply,zeng2020soft}, which automatically learn the effective image representations, have gained many successes in facial age estimation. 

However, the most promising methods~\cite{niu2016ordinal,tan_pami_2018,tan2019deeply,carletti2019age,punyani2020neural,othmani2020age} use ensembles of DCNNs, making the obtained classifier not usable in real applications. These methods usually require prohibitive computational resources, complex training procedures caused by the plurality of neural networks, and substantial training sets, which are not simply collectible.

Motivated by this concern, we build a strong baseline based on a single neural network by collecting some efficient tricks in pre-training, model architecture, data augmentation, and so on. The strong baseline achieves impressive performance compared to the standard baseline. Nevertheless, the resulting estimator is still not a balance age estimator and usually shows poor performance in classes with fewer samples.

Unfortunately, existing methods for age estimation~\cite{niu2016ordinal,gao2018age,zeng2020soft} still lack in long-tailed scenarios. The long-tail distribution of the visual world poses great challenges for deep learning-based classification models on how to handle the data imbalance problem. Inspired by the existing work~\cite{kang2019decoupling}, we decouple the learning procedure into representation learning and classification to train a balanced estimator.

In the representation learning stage, we obtain a robust feature extractor to distinguish different ages. Furthermore, in the classification stage, we freeze the model except for the FC layer and train the model on balanced MIVIA~\cite{greco2021effective} dataset obtained by class-balanced sampling~\cite{shen2016relay}. Moreover, we innovatively introduce Mean Square Error (MSE) loss into this stage for further narrowing down the standard deviation of different age groups. By applying this, we obtain a model that performs well at all ages.

To summarize, the main contributions are listed as follows:
\begin{itemize}

\item We collect some effective training tricks to build a strong baseline based on a single neural network, which effectively enhances age estimation performance.

\item We propose a novel LAE method to deal with the age estimation in long-tailed scenarios. The resulting estimator performs well at all ages.

\item Extensive experiments on MIVIA indicate the effectiveness of LAE.
\end{itemize}

\section{Proposed Method}

\subsection{Overview}
In our implementation, we break the neural network into two components, a feature extractor $\theta$ and a classifier $\psi$. The classifier consists of a linear layer and a softmax layer. Before explaining the proposed method in detail, we first define the key symbols.
Assume we have an age dataset $D$. The $i$th sample in the $D$ is denoted by $(x_i, y_i, z_i)$, where $x_i$ denote the input image, $y_i$ denote the age label, and $z_i$ denote the label distribution. The label distribution of the sample is based on a typical Gaussian distribution, \ie, $ z_{i}^k = \frac{1}{\sqrt{2\pi}\sigma} exp{(-\frac{(k-y_{i})^2}{2\sigma^2})}$ where $k \in [0, ..., 100]$ and standard deviation $\sigma$ is set to 1.

\subsection{Standard Baseline}
In this subsection, we provide a standard baseline used for comparison. During the training stage, the pipeline includes the following steps:

1. We initialize the ResNet18~\cite{he2016deep} with pre-trained parameters on ImageNet~\cite{deng2009imagenet} and change the dimension of the fully connected layer to K. And K is set to 101 in our experiments.

2. We resize each image into 224 × 224 pixels and each image is flipped horizontally with 0.5 probability.

3. Each image is decoded into 32-bit floating point raw pixel values in [0, 1]. Then we normalize RGB channels by subtracting 0.485, 0.456, 0.406 and dividing by 0.229, 0.224, 0.225, respectively.



4. Following the work~\cite{gao2018age}, the network is trained with a KL divergence and $\ell_1$ loss. KL divergence is calculated by measuring the distance between label distribution $z_i$ and predicted age distribution $\hat{z}_i$, and $\ell_1$ loss aims to narrow the distance between the predicted age $\hat{y}_i$ and the ground truth label $y_i$. Specifically, the predicted age $\hat{y}_i$ is obtained by using an expectation regression on the softmax outputs.

5. SGD is adopted to optimize the model. The initial learning rate is set to be 0.005 and is decreased by 0.1 at the 20$th$, 40$th$, and 60$th$ epoch. Totally there are 75 epochs.

\subsection{Training Tricks}
In this subsection, we introduce some effective tricks for training. After expanding all the tricks, the standard baseline becomes a strong baseline.

\paragraph{Model Architecture}:
Compared to the previous models, EfficientNetV2~\cite{tan2021efficientnetv2} has a faster training speed and better parameter efficiency. In our submission, we use the EfficientNetV2-M as the backbone of the feature extractor.

\paragraph{Pre-training: } Pre-training has a significant impact on the performance of an age estimator. Before training the network on age datasets, we pretrain the model on MS-celeb-1M~\cite{guo2016ms}, with only using identity labels. Following the work~\cite{wang2018cosface}, which AM-softmax loss is employed as the loss function. In this way, a large number of face images are employed to enhance the network's discriminating ability with obtaining a good initialization.

\paragraph{Batch Size: } The batch size also shows its impact on performance during training. Compared to 16, 64, and 128, keep 32 samples in one GPU shows the best performance. So the default batch size for us is 256 since we use 8 GPUs for training.

\paragraph{Onecycle Scheduler: } Existing methods~\cite{tan_pami_2018,gao2018age} for age estimation usually require many epochs to convergence, which undoubtedly brings a more significant time cost and occupation of computing resources. The existence of Onecycle~\cite{smith2019super} can make networks be trained much faster than before. In our case, it reduces the number of training epochs from the original 75 to 24.

\paragraph{Data Augmentation: } 
As a data-space solution to the problem of limited data, data augmentation is capable of enhancing the size and quality of training datasets to avoid overfitting and make the resulting model more generalization. In our case, we applied data augmentation as follows: first, a random patch of the images is selected and resized to 224 × 224 with a random horizontal flip, followed by a color jitter, consisting of a random sequence of brightness, contrast, saturation, hue adjustments. Finally, RandAugment~\cite{cubuk2020randaugment} is applied, and we set N=2 and M=9, where N denotes the number of transformations to apply, and M denotes the magnitude of the applied transformations.

\begin{figure}[h]
\caption{The distribution of the dataset samples over the age is depicted on the left. The long-tailed distribution of the MIVIA is depicted on the right, sorted by the number of samples in class. Head-class denotes high-frequency class, and tail-class denotes low-frequency class. In practice, the number of samples per class generally decreases from head to tail classes.
}
\includegraphics[width = \linewidth]{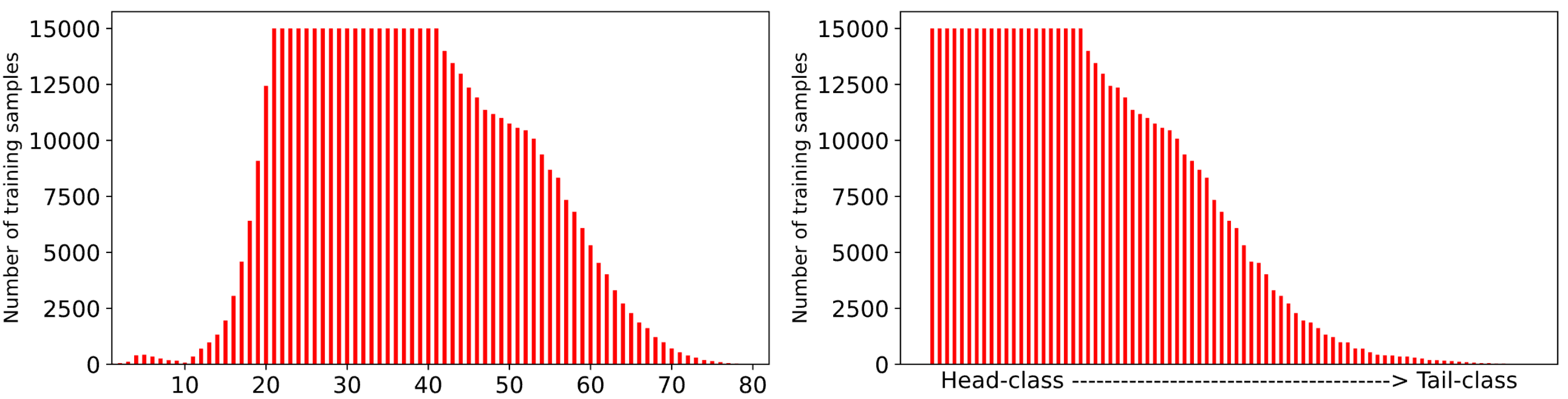}
\label{fig:1}
\end{figure}

\subsection{Long-tailed Age Estimation}
The long-tail distribution of the visual world poses great challenges for deep learning-based classification models on how to handle the data imbalance problem~\cite{kang2019decoupling,ren2020balanced}. As the result, the resulting model usually shows poor performance in classes with fewer samples. This phenomenon is reflected in most of the age datasets. As shown in Fig.~\ref{fig:1}, the age distribution of the MIVIA is imbalanced, in line with the premise of long-tailed recognition. Inspired by related research in the field of long-tailed recognition~\cite{kang2019decoupling}, we decouple the learning procedure into representation learning and classification to train a relatively balanced classifier, which performs well at all ages. 

\paragraph{Representation Learning Stage:}
 In this stage, we follow the pipeline of strong baseline for training a robust feature extractor to distinguish different ages. To this end, we finetune the model on the MIVIA training set. By feeding the feature maps into the classifier, we obtain the prediction distribution $\hat{z}^k_{i}$, which indicates the probability of classifying $x_i$  to age $k$. Then, we use KL divergence to measure the label distribution loss $L_{ld}$ between ground-truth label distribution $z_{i}^k = \frac{1}{\sqrt{2\pi}\sigma} exp{(-\frac{(k-y_{i})^2}{2\sigma^2})}$ and the prediction distribution $\hat{z}^k_{i}$~\cite{gao2018age}:

\begin{equation}
    L_{ld} = \sum_{k=1}^K  z^k_{i}  log \frac{z^k_{i}}{\hat{z}^k_{i}} 
\label{L_sr}
\end{equation}%

The predicted age $\hat{y}_i$ is obtained by using a expectation regression\cite{rothe2015dex} on the output distribution. It can be denoted as $\hat{y}_i=\sum_{k=1}^K k \hat{z}^k_{i}$. Moreover, we employ a loss $L_{er}$ to further narrowing the distance between the predicted age label and the ground-truth label :

\begin{equation}
    L_{er} = | y_{i} - \sum_{k=1}^K k \hat{z}^k_{i}|  
\label{L_er1}
\end{equation}%
where $| \cdot |$ denotes $\ell_1$ loss.

Finally, we integrate all the above loss functions and train them simultaneously. Referring to the losses in Eq.~\ref{L_sr}, ~\ref{L_er1} as $L_{ld}, L_{er}$, respectively, then overall we optimize the feature extractor with $L_{fe}$: 

\begin{equation}
    L_{fe} = L_{ld} + \lambda L_{er}
\end{equation}%
where $\lambda$ denotes the weight of $L_{er}$.

\paragraph{Classification Stage:} 
To retrain the classifier without changing the parameters of the feature extractor, we freeze the model except for the FC layer and train on balanced MIVIA obtained by class-balanced sampling, which means each class has an equal probability of being selected. One can see this as a two-stage sampling strategy, where a class is first selected uniformly from the set of classes, and then an instance from that class is subsequently uniformly sampled~\cite{shen2016relay}.

Moreover, we innovatively introduce MSE loss into this stage for further narrowing down the standard deviation of different age groups. To this end, we optimize the classifier with $L_{cl}$:

\begin{equation}
    L_{cl} = L_{ld} + (L_{er} - mae)^2
\end{equation}%
where $mae$ denotes the mean absolute error over the validation set.

\begin{equation}
    mae = \sum_{i=1}^M \frac{| y_{i} - \hat{y_i}| }{M} 
\label{L_er}
\end{equation}%
where $M$ is the total number of validating images, $y_i$ and $\hat{y_i}$ denote the ground-truth age and the predicted age, respectively. It is worth noting that the $mae$ here results from the representation learning stage on the validation set.

\section{Experiments}

In this section, we first introduce the datasets and the evaluation metrics from our experiments. Then we provide implementation details for our LAE method, describe experimental setups, and discuss results.

\subsection{Dataset}

The MIVIA Age Dataset is composed of 575,073 images of more than 9,000 identities. Those images have been extracted from the VGGFace2~\cite{cao2018vggface2} dataset and annotated with age by means of a knowledge distillation technique, making the dataset very heterogeneous in terms of face size, illumination conditions, facial pose, gender, and ethnicity.

\subsection{Metric}

An index called Age Accuracy and Regularity (AAR) is introduced for taking into account accuracy and regularity:

\begin{equation}
    AAR = max(0; 7-MAE) + max(0; 3-\sigma) 
\label{L_er}
\end{equation}%
where MAE denotes the mean absolute error on the entire test set, and $\sigma$ is obtained by:

\begin{equation}
    \sigma =  \sqrt{\frac{\sum^8_{j=1}(MAE^j - MAE)^2}{8}}
\label{L_er}
\end{equation}%
where $MAE^j$ denotes the mae and is computed over the samples whose real age is in age group $j$. The details of age group are shown as below.

 \setlength{\tabcolsep}{10pt}
\begin{table}[h]
\centering
\caption{The details of the eight age groups.}
\resizebox{\linewidth}{!}{
\begin{tabular}{cccccccc}
\toprule[1pt]
 \textbf{$MAE^1$}   & \textbf{$MAE^2$} & \textbf{$MAE^3$} & \textbf{$MAE^4$} & \textbf{$MAE^5$} & \textbf{$MAE^6$} & \textbf{$MAE^7$} & \textbf{$MAE^8$}\\ \midrule[1pt]
 1-10      & 11-20  & 21-30 &31-40 &41-50 &51-60 &61-70 &71-81    \\
 \bottomrule[1pt]
\end{tabular}
}
\label{tab:age group}
\end{table}

\subsection{Implementation Details}


We use an SGD optimizer with a MultiStepLR scheduler over 18 epochs in the pre-training stage, with the milestones of [7, 14, 16].
Moreover, we use an SGD optimizer with a OneCycle schedule over 24 epochs and 8 epochs in the representation and classification stage. We set the base learning rate to 0.005 and 0.001, scaled linearly with the batch size (LearningRate = base × BatchSize/256). In addition, we use a global weight decay parameter of  5e-4. The default batch size is set to 256. According to the work~\cite{zhang2016joint}, the images are aligned with five landmarks (including two eyes, nose tip, and two mouth corners). For a fair comparison, we fix the $\lambda = 1$ during all experiments.

\begin{figure}[h]
\caption{Examples of apparent age estimation using LAE on MIVIA testing images.}

\includegraphics[width = \linewidth]{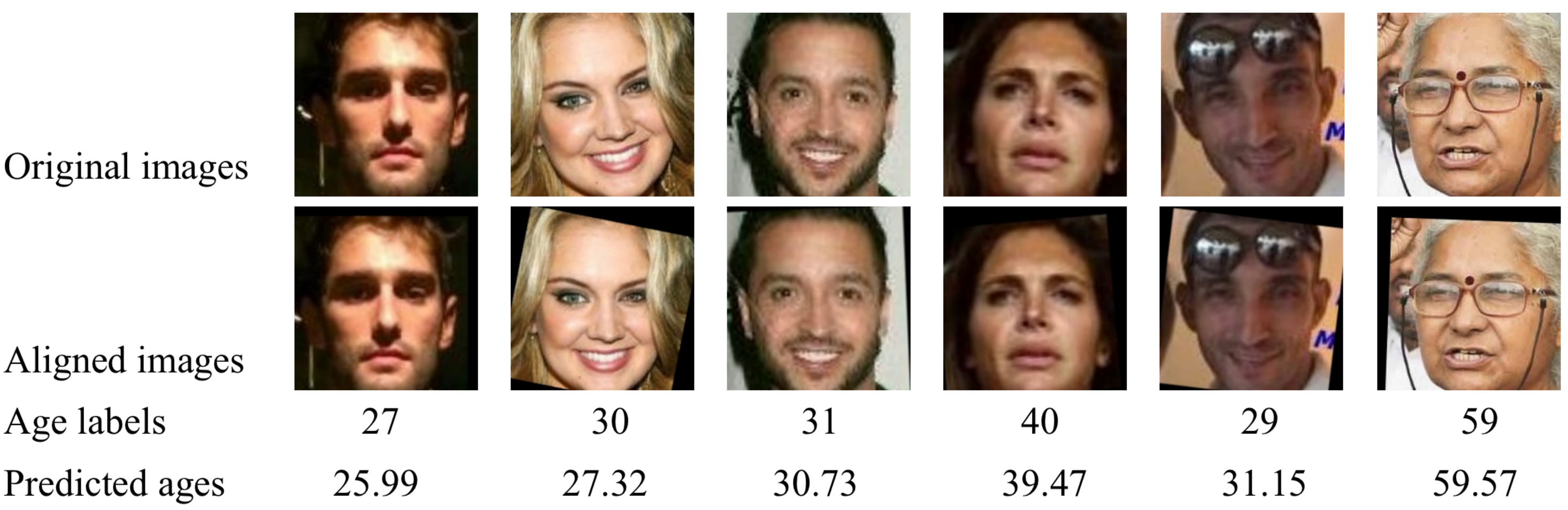}
\vspace{-0.7cm}
\label{fig:2}
\end{figure}

\subsection{Experimental Results}
Since the test set is not available during the competition, we randomly divide the MIVIA into a training set and a validation set at a ratio of 4:1 to validate the effectiveness of the proposed methods. The result is shown as below. The standard baseline achieves 1.97 on MAE. Then, we add model architecture, pre-training, and data augmentation, respectively. The final strong baseline boosts more performance than other models. Finally, the bag of tricks makes the estimator achieves 1.71 on MAE and 7.18 on AAR. The results (such as in Fig. 2) shows that our proposed solution is able to predict the age of faces as well as people can or even better.

Moreover, the strong baseline is also the result of our representation learning stage. The estimator performs well at head-classes but shows poor performance at tail-classes. As a result, it also performs poorly on the $\sigma$ and the AAR. Then, we add the classification stage into the training procedure, and the final estimator shows the best performance on tail-classes, $\sigma$, and AAR on the MIVIA. 

The results on test sets are provided by the organizers. And it is worth noting that the submission is trained on the whole MIVIA. 

 \setlength{\tabcolsep}{8pt}
\begin{table}[h]
\centering
\caption{The performance of different models is evaluated on MIVIA. Baseline-S stands for the standard baseline introduced in Section \textcolor{red}{2.2}. $MAE^1$ and $MAE^8$ stand for age groups consisting of tail-classes, and $MAE^3$ stands for head-classes. $*$ denotes the model is evaluated on the test set, and other models are all evaluated on the validation set. The best results on the validation set are in bold.}
\resizebox{\linewidth}{!}{
\begin{tabular}{l|cccc|cc}
\toprule[1pt]
Model               & \textbf{$MAE$} & \textbf{$MAE^1$} & \textbf{$MAE^3$} & \textbf{$MAE^8$} & \textbf{$\sigma$} & \textbf{$AAR$} \\ \midrule[0.5pt]
Baseline-S          & 1.97    & 7.16    & 1.61    & 3.45    &  1.93     & 6.10    \\
+model architecture &  1.93   &  6.47   & 1.59    & 2.99    &  1.65     & 6.42    \\
+pre-training       &  1.78   & 5.09    & 1.45    & 2.73    &  1.33     & 6.89    \\
+data augmentation        & 1.85    & 5.53    & 1.52    & 2.99    & 1.37      & 6.78    \\ \midrule[0.5pt]
Representation (Strong Baseline)          & \textbf{1.71}    & 4.66    & \textbf{1.40}    & 2.59    & 1.11      & 7.18    \\
Classification  & 1.89    & \textbf{2.69}    & 1.62    & \textbf{2.11}    &  \textbf{0.37}     & \textbf{7.74}    \\\midrule[0.5pt]
Submission$*$       & 1.86    &  /   &  /   &  /   &  0.20     & 7.94    \\
 \bottomrule[1pt]
\end{tabular}
}
\label{tab:trick}
\end{table}

\section{Conclusions}
In this paper, we first formulate a standard baseline and build a strong baseline by collecting the tricks in pre-training, model architecture, data augmentation, and so on. The bag of tricks significantly enhances the age estimation performance. The resulting estimator achieves the best performance on head-classes. Second, we propose an LAE method, which decouples the learning procedure into representation learning and classification, to obtain a more balanced estimator. The proposed method performs well at all ages, significantly outperforming the human reference.

\section*{Acknowledgments}
This work was supported by the Chinese National Natural Science Foundation Projects $\#$61961160704, $\#$61876179, the External cooperation key project of Chinese Academy Sciences $\#$ 173211KYSB20200002, the Key Project of the General Logistics Department Grant No.AWS17J001, Science and Technology Development Fund of Macau (No.0010/2019/AFJ, 0008/2019/A1, 0025/2019/A- \\ KP, 0019/2018/ASC).

\bibliographystyle{splncs04}

\end{document}